\documentclass[10pt,twocolumn,letterpaper]{article}

\usepackage{cvpr}
\usepackage{times}
\usepackage{epsfig}
\usepackage{graphicx}
\usepackage{amsmath}
\usepackage{amssymb}
\usepackage{algorithm}
\usepackage[noend]{algpseudocode}

\usepackage[breaklinks=true,bookmarks=false]{hyperref}

\cvprfinalcopy 

\def\cvprPaperID{15} 

\begin{document}

\title{Accelerated CNN Training Through Gradient Approximation}

\author{Ziheng Wang\\
Department of EECS, M.I.T\\
77 Massachusetts Ave, Cambridge, MA\\
{\tt\small ziheng@mit.edu} 
\and
Sree Harsha Nelaturu\\
Department of ECE, SRMIST\\
Kattankulathur, Chennai, TN\\
{\tt\small sreeharsha\_murali@srmuniv.edu.in}}


\maketitle

\begin{abstract}

Training deep convolutional neural networks such as VGG and ResNet by gradient descent is an expensive exercise requiring specialized hardware such as GPUs. Recent works have examined the possibility of approximating the gradient computation while maintaining the same convergence properties. While promising, the approximations only work on relatively small datasets such as MNIST. They also fail to achieve real wall-clock speedups due to lack of efficient GPU implementations of the proposed approximation methods. In this work, we explore three alternative methods to approximate gradients, with an efficient GPU kernel implementation for one of them. We achieve wall-clock speedup with ResNet-20 and VGG-19 on the CIFAR-10 dataset upwards of 7 percent, with a minimal loss in validation accuracy. 

\end{abstract}

\section{Introduction}

Deep convolutional neural networks (CNN) are now arguably the most popular computer vision algorithms. Models such as VGG \cite{vgg} and ResNet \cite{he2016deep} are widely used. However, these models contain up to hundreds of millions of parameters, resulting in high memory footprint, long inference time and even longer training time. 

The memory footprint and inference time of deep CNNs directly translate to application size and latency in production. Popular techniques based on model sparsification are able to deliver orders of magnitude reduction in the number of parameters in the network. \cite{han2015deep} Together with emerging efficient sparse convolution kernel implementations, deep CNNs can now be realistically used in production after training \cite{gray2017gpu, park2016faster,chen2018escort}. 

However, the training of these deep CNNs is still a lengthy and expensive process. The hundreds of millions of parameters in the model must all be iteratively updated hundreds of thousands of times in a typical training process based on back-propagation. Recent research has attempted to address the training time issue by demonstrating effective training on large scale computing clusters consisting of thousands of GPUs or high-end CPUs \cite{you2018imagenet,akiba2017extremely,jia2018highly}. However, these computing clusters are still extremely expensive and labor-intensive to set up or maintain, even if the actual training process is reduced to minutes. 

An alternative to using large computing clusters is to accelerate the computations of the gradients themselves. One option is to introduce highly optimized software \cite{chetlur2014cudnn} or new hardware  \cite{markidis2018nvidia, jouppi2017datacenter}. The training can also be performed in lower precision, which can lead to massive speedups with appropriate hardware support \cite{micikevicius2017mixed}. Another less pursued option, complementary to the previous two, is to approximate the actual gradient computation themselves \cite{sun2017meprop, sun2018training, wei2017minimal,adelman2018faster}. Other recent works have also suggested that the exact gradient might not be necessary for efficient training of deep neural networks. Studies have shown that only the sign of the gradient is necessary for efficient back propagation  \cite{xiao2018biologically,wen2017terngrad}. Surprisingly, even random gradients can be used to efficiently train neural networks  \cite{lillicrap2016random,nokland2016direct}. However, these findings are mostly limited to small fully connected networks on smaller datasets. The approximation algorithms proposed also cannot directly translate into real wall-clock speedups in training time due to lack of efficient GPU implementation.

In this work, we hypothesize that we can extend gradient approximation methods to deep neural networks to speed up gradient computations in the training process. We hypothesize that we can apply these approximations to only a subset of the layers and maintain the validation accuracy of the trained network. We validate our hypotheses on three deep CNNs (2-layer CNN \cite{krizhevsky2009learning}, ResNet-20  \cite{he2016deep} VGG-19 \cite{vgg}) on CIFAR-10. Our methods are fully compatible with classic deep CNN architectures and do not rely on explicit sparsity information that must be input to the network, like approaches such as SBnet and Sub-manifold networks \cite{ren2018sbnet,graham2017submanifold}.  

We summarize our contributions as follows: 
\begin{itemize}
  \item We present three gradient approximation methods for training deep CNNs, along with an efficient GPU implementations for one of them. 
  \item We explore the application of these methods to deep CNNs and show that they allow for training convergence with minimal validation accuracy loss.
  \item We describe the concept of approximation schedules, a way to reason about applying different approximation methods across different layers and training batches.
\end{itemize}

\section{Approximation Methods}

\begin{figure}[t]
\begin{center}
\includegraphics[width=0.6\linewidth]{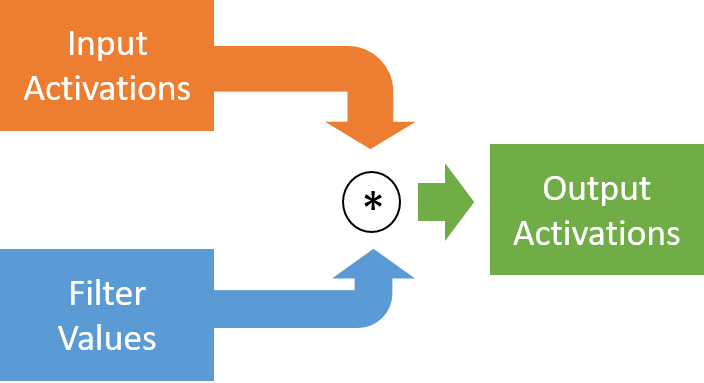}
\includegraphics[width=0.8\linewidth]{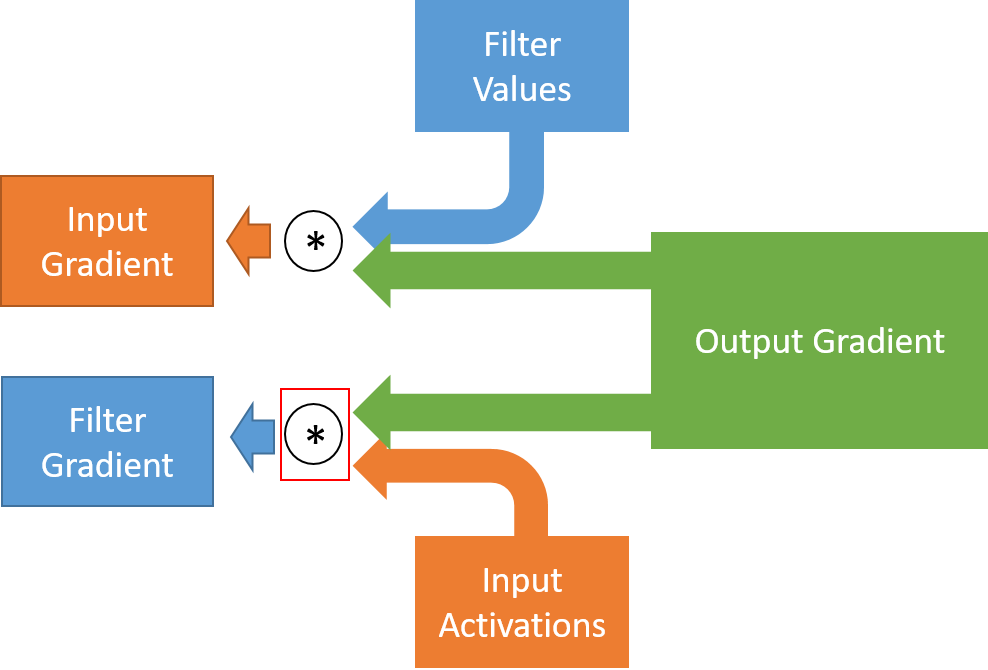}
\end{center}
   \caption{Forward and backward propagation through a convolutional layer during training. Asterisks indicate convolution operations and the operation in the red box is the one we approximate.}
\label{fig:back}
\label{fig:onecol}
\end{figure}

In a forward-backward pass of a deep CNN during training, a convolutional layer requires three convolution operations: one for forward propagation and two for backward propagation, as demonstrated in Figure \ref{fig:back}. We approximate the convolution operation which calculates the gradients of the filter values, which constitutes roughly a third of the computational time. We aim to apply the approximation a quarter of the time across layers/batches. This leads to a theoretical maximum speedup of around 8 percent. 

\subsection{Zero Gradient}
The first method passes back zero as the weight gradient of a chosen layer for a chosen batch. If done for every training batch, it effectively freezes the filter weights.  

\subsection{Random Gradient}
The second method passes back random numbers sampled from a normal distribution with mean 0 and standard deviation $\frac{1}{128}$ (inverse of batch size) as the weight gradient of a chosen layer for a chosen batch. Different values in the weight gradient are chosen independently. Importantly, this is different from the random feedback alignment method discussed in \cite{lillicrap2016random} and \cite{nokland2016direct} as we regenerate the random numbers every training batch. We implement this using tf.py\_func, where np.random.normal is used to generate the random values. This approach is extremely inefficient, though surprisingly faster than a naive cuRAND implementation in a custom tensorflow operation for most input cases. We are working on a more efficient implementation. 

\subsection{Approximated Gradient}
The third method we employ is based on the top-k selection algorithms popular in literature. \cite{wei2017minimal} In the gradient computation for a filter in a convolutional layer, only the largest-magnitude gradient value is retained for each output channel and each batch element. They are scaled according to the sum of the gradients in their respective output channels so that the gradient estimate is unbiased, similar to the approach employed in \cite{wangni2018gradient}. All other gradients are set to zero. This results in a sparsity ratio of $1-\frac{1}{HW}$, where $H$ and $W$ are the height and width of the output hidden layer. The filter gradient is then calculated from this sparse version of the output gradient tensor with the saved input activations from the forward pass. The algorithm can be trivially modified to admit the top-k magnitude gradient values with an adjustment of the scaling parameter, a direction of future research. Similar to the random gradient method, we find that we need to scale our approximated gradient by a factor proportional to the batch size for effective training. In the experiments here, we scale them by $\frac{1}{128}$.

\subsection{Efficient GPU Implementation}

A major contribution of this work is an implementation of the approximated gradient method in CUDA. This is critical to achieve actual wall-clock training speedups. A naive Tensorflow implementation using tf.image.extract\_glimpse does not use the GPU and results in significantly slower training time. 

\begin{figure}[t]
\begin{center}
\includegraphics[width=1\linewidth]{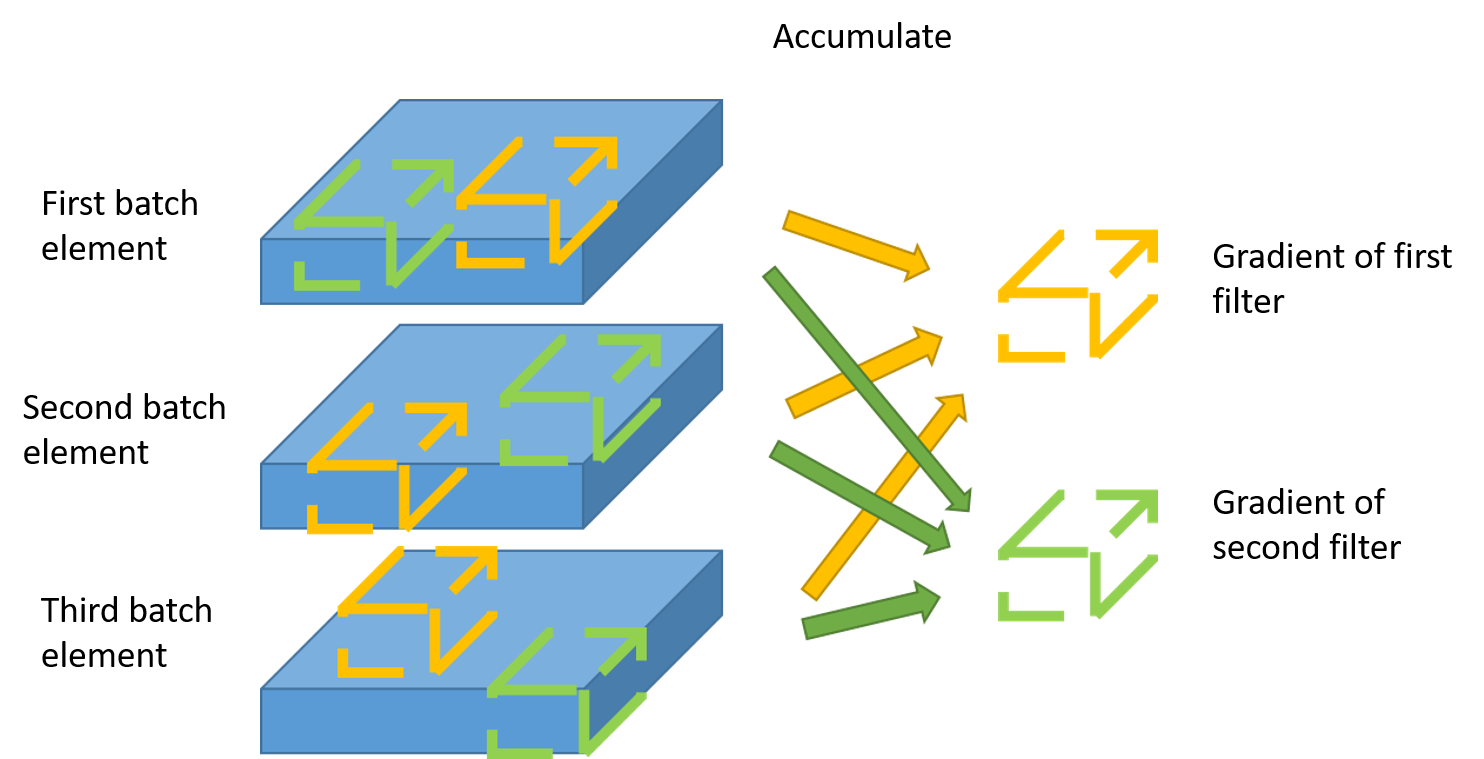}
\end{center}
   \caption{The approximation algorithm illustrated for an example with two filters and three input elements. For each filter, we extract a patch from each batch element's input activations and accumulate the patches.}
\label{fig:algo}
\end{figure}

Efficient GPU implementations for dense convolutions frequently use matrix lowering or transforms such as FFT or Winograd \cite{chetlur2014cudnn,liu2018efficient}.  However, the overheads of these transformations might not be worth the benefit in a sparse setting. Recent approaches have sought to perform the sparse convolution directly on CPU or GPU \cite{park2016faster,chen2018escort}. Here we also opt for the latter approach. We interpret the sparse convolution in the calculation of the filter gradient as a patch extraction procedure, as demonstrated in Figure \ref{fig:algo}.

Formally, let's assume we have an input tensor $I$ and output gradient tensor $dO$ in $NCHW$ format, where $N$ is the batch dimension, $C$ the channel dimension and $H$, $W$ the height and width of the hidden layer. The filter tensor, $f$, has dimension $KKC_iC_o$, where $K$ is the filter size, $C_i$ is the number of channels in $I$ and $C_o$ is the number of channels in $dO$. We will use the symbol $\ast$ to denote the convolution operation. In order to compute $df$, we have to convolve $I$ with $dO$. If we zero out all elements in $dO$ except one for each output channel dimension, then convolution becomes a collection of $C_o$ patches of shape $KKC_i$ from $I$, as specified below in Algorithm 1. 

\begin{algorithm}
\caption{Max Gradient Approximation}\label{euclid}
\begin{algorithmic}[1]

\State $df[:,:,:,:] = 0$
\For {$c = 1:C_o$}
\For {$n = 1:N$}
\State $row, col\gets \arg\max abs(dO[n,c,:,:])$ 
\State $sum \gets \sum dO[n,c,:,:]$ (sum is a scalar)

\State $df[:,c,:,:] += I[n,:,row:row+K,col:col+K] * sum$
\EndFor
\EndFor

\end{algorithmic}
\end{algorithm}

Our kernel implementation expects input activations in $NHWC_i$ format and output gradients in $NC_oHW$ format. It produces the output gradient in $C_oKKC_i$ format. In $NHWC$ format, GPU global memory accesses from the patch extractions can be efficiently coalesced across the channel dimension, which is typically a multiple of 8. Each thread block is assigned to process several batch elements for a fixed output channel. Each thread block first computes the indices and values of the nonzero weight values from the output gradients. Then, they extract the corresponding patches from the input activations and accumulate them to the result. 

We benchmark the performance of our code against NVIDIA cuDNN v7.4.2 library apis. Approaches such as cuSPARSE have been demonstrated to be less effective in a sparse convolution setting and are not pursued here \cite{chen2018escort}. All timing metrics are obtained on a workstation with a Titan-Xp GPU and 8 Intel Xeon CPUs at 3.60GHz.

All training experiments are conducted in $NCHW$  format, the preferred data layout of cuDNN. As a result, we incur a data transpose overhead of the input activations from $NCHW$ to $NHWC$. In addition, we also incur a slight data transpose overhead of the filter gradient from $C_oKKC_i$ to $KKC_iC_o$. 

\subsection{Approximation Schedules}

\begin{figure}[t]
\begin{center}
\includegraphics[width=1\linewidth]{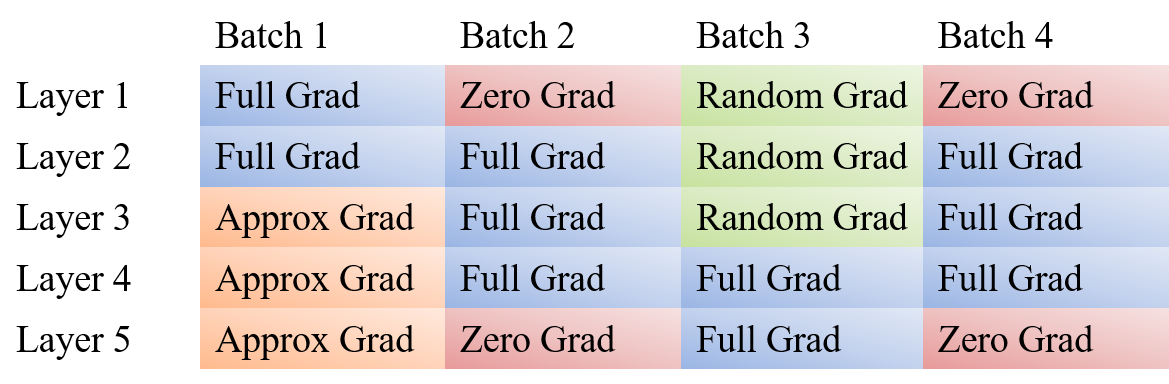}
\end{center}
   \caption{Example approximation schedule for a 5-layer network over 4 training batches. Full Grad denotes regular gradient computation without approximation.}
\label{fig:schedule}
\end{figure}

Here, we introduce the concept of approximation schedules. This concept allows us to specify when particular approximations are applied in the training process and how to combine different approximations. Existing approximation methods such as DropBack \cite{golub2018dropback} and PruneTrain \cite{lym2019prunetrain} can be applied to a specific layer, but are applied over all training batches since their application at a particular training batch changes the structure of the network, thus affecting all subsequent batches. The three approximation methods we have mentioned approximate the gradient computation of a weight filter for a single training batch. They can be thus applied to a specific layer for a specific training batch. We refer to the term "approximation schedule" as a specification of what approximation method to apply for each layer and each training batch that is consistent with the above rules. An example of an approximation schedule is shown in Figure \ref{fig:schedule}. More aggressive approximation schedules might lead to a higher loss in accuracy, but would also result in higher speedups. Here, we demonstrate that simple heuristics to pick approximation schedules can lead to good results on common networks such as ResNet-20 and VGG-19. While the efficacy of simple heuristics is crucial for the applicability of the proposed approximation methods in practice, determining the optimal approximation schedule for different neural network architectures is an interesting direction of future research.

\section{Evaluation}
\begin{figure}[t]
\begin{center}
\includegraphics[width=0.7\linewidth]{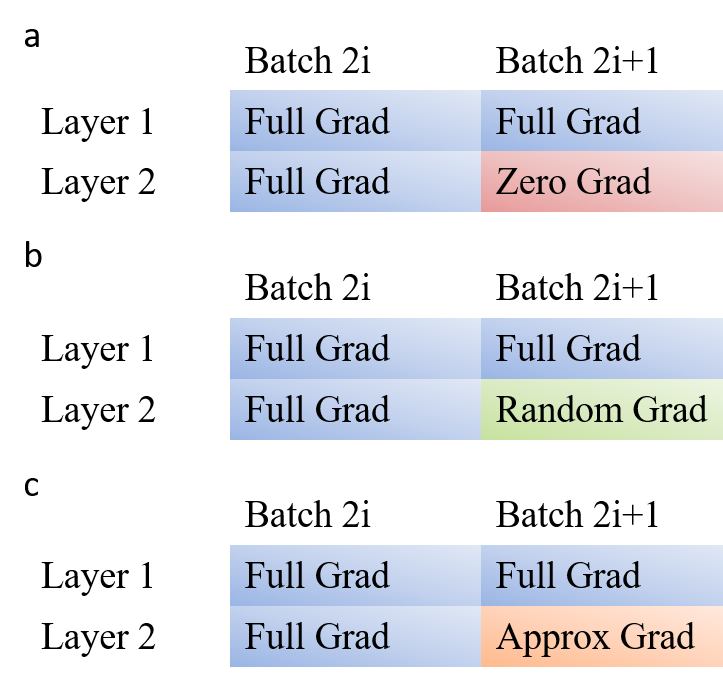}
\end{center}
   \caption{The three approximation schedules studied for the 2-layer network using a) zero gradient method b) random gradient method c) approximated gradient method.}
\label{fig:2-layer-schedule}
\end{figure}

We test our approach on three common neural network architectures (2-layer CNN \cite{krizhevsky2009learning}, VGG-19 \cite{vgg} and ResNet-20 \cite{he2016deep}) on the CIFAR-10 dataset. The local response normalization in the 2-layer CNN is replaced by the more modern batch normalization method \cite{ioffe2015batch}. For all three networks, we aim to use the approximation methods 25 percent of the time. In this work, we test all three approximation methods separately and do not combine. On the 2-layer CNN, we apply the selected approximation method to the second convolutional layer every other training batch. On VGG-19 and ResNet-20, we apply the selected approximation method to every fourth convolutional layer every training batch, starting from the second convolutional layer. For example, the three approximation schedules for the 2-layer CNN are shown in Figure \ref{fig:2-layer-schedule}. We start from the second layer because recent work has shown that approximating the first convolutional layer is difficult \cite{adelman2018faster}. This results in four approximated layers for VGG-19 and five approximated layers for ResNet-20. For the ResNet-20 model, we train a baseline ResNet-14 model as well. Training a smaller model is typically done in practice when training time is of concern. Ideally, our approximation methods to train the larger ResNet-20 model should result in higher validation accuracy than the ResNet-14 model. For ResNet-20, we also experiment with other approximation schedules to show that our approximation methods are robust to schedule choice. 

We train the networks until the validation accuracy stabilizes. It took around 500 epochs for the 2-layer CNN, 250 epochs for the ResNet-20 model, and 200 epochs for the VGG-19 model. We use an exponentially decaying learning rate with the Adam optimizer for all three models. We apply typical data augmentation techniques such as random cropping and flipping to each minibatch. All training was performed within Tensorflow 1.13 \cite{abadi2016tensorflow}. 

\subsection{Performance Comparisons}

We compare the performance of our GPU kernel for the approximated gradient method with the full gradient computation for the weight filter as implemented in cuDNN v7.4.2. cuDNN offers state-of-the-art performance in dense gradient computation and is used in almost every deep learning library. For each input case, cuDNN tests several hand-assembled kernels to pick the fastest one. The kernels fully utilize the high floating point throughput of the GPU to perform the dense gradient computations. In contrast, sparse approximations of the gradient usually involve less arithmetic/memory ratio and do not admit as efficient kernel implementations on GPU. It is oftentimes necessary to impose structure or high sparsity ratio to achieve actual performance gain \cite{zhu2018structurally}. Here we demonstrate that our gradient approximation method does yield an efficient GPU implementation that can lead to actual speedups compared to cuDNN. 

We present timing comparisons for a few select input cases encountered in the network architectures used in this work in Table \ref{table:perf}. We aggregate the two data transpose overheads of the input activations and the filter gradients. (In almost every case, the data transpose overhead of the input activations dominates.) We make three observations. 

Firstly, in most cases, the gradient approximation, including data transposition, is at least three times as fast as the cuDNN baseline. Secondly, we observe that cuDNN timing scales with the number of input channels times the height and width of the hidden layer, whereas our approximation kernel timing scales with the number of input channels alone. This is expected from the nature of the computations involved: the performance bottleneck of our kernel is the memory intensive patch extractions, the sizes of which scale with the number of input channels times filter size. Thirdly, we observe that in many cases, the data transposition overhead is over fifty percent of the kernel time, suggesting that our implementation can be further improved by fusing the data transpose into the kernel as in SBNet \cite{ren2018sbnet}. This is left for future work. 

\begin{table}[t]
\begin{center}
\includegraphics[width=1\linewidth]{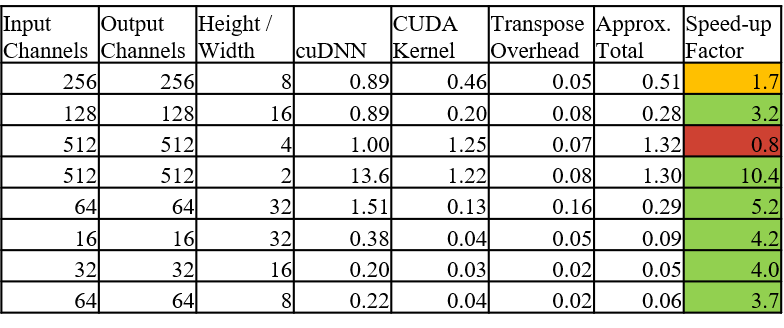}
\end{center}
   \caption{Performance comparisons. All timing statistics in microseconds. Approx. total column is the sum of the CUDA Kernel time and the transpose overhead.}
\label{table:perf}
\end{table}

\subsection{Training Convergence}

\begin{figure}[t]
\begin{center}
\includegraphics[width=1\linewidth]{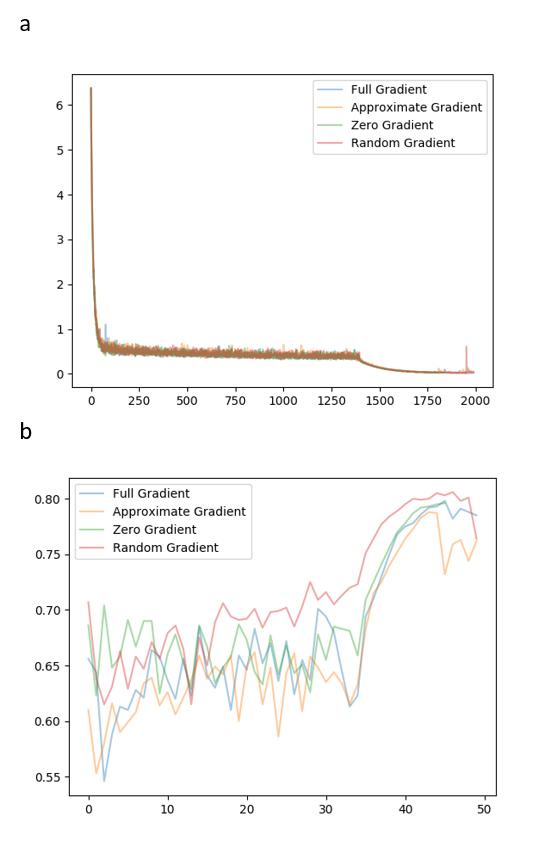}
\end{center}
   \caption{a) Training loss of 2-layer CNN with different approximation methods. b) Validation accuracy of 2-layer CNN with different approximation methods.}
\label{fig:2acc}
\end{figure}

We present convergence results for the training of our three neural networks using the chosen approximation schedules with two metrics, training loss and validation accuracy. In Figure \ref{fig:2acc}, we see that for the 2-layer CNN, all approximation methods result in training loss curves and validation accuracy curves similar to the ones obtained by full gradient computation. We can even see that the random gradient method surpasses full gradient computation in terms of validation accuracy. The zero gradient method is very similar to full gradient computation while the approximated gradient method does slightly worse. 

\begin{figure}[t]
\begin{center}
\includegraphics[width=1\linewidth]{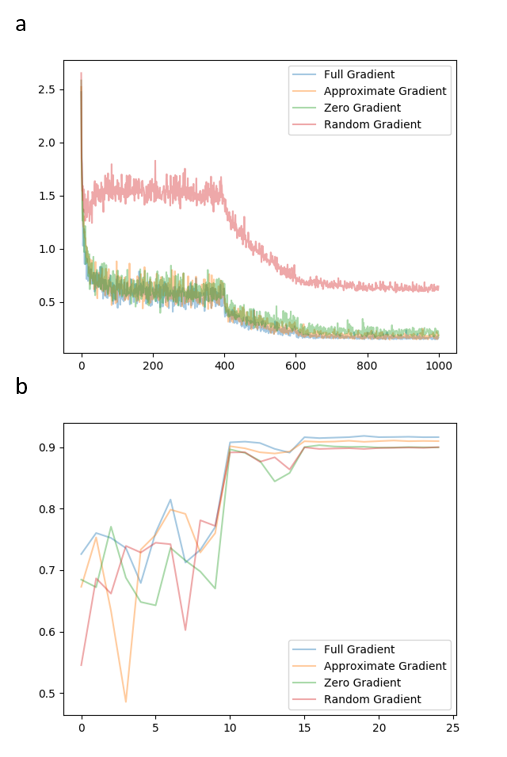}
\end{center}
   \caption{a) Training loss of ResNet-20 with different approximation methods. b) Validation accuracy of ResNet-20 with different approximation methods. The loss curve of the random gradient method stagnates but the validation accuracy is competitive.}
\label{fig:racc}
\end{figure}

The approximation methods remain robust on larger networks, such as ResNet-20, shown in Figure \ref{fig:racc}. In this case, we can see from both the loss curves and the validation accuracy that our approximated gradient methods do slightly worse than full gradient computation, but better than both random gradient or zero gradient methods. Curiously, the random gradient method maintains a high training loss throughout the training, but is still able to achieve good validation accuracy. 

\begin{figure}[t]
\begin{center}
\includegraphics[width=1\linewidth]{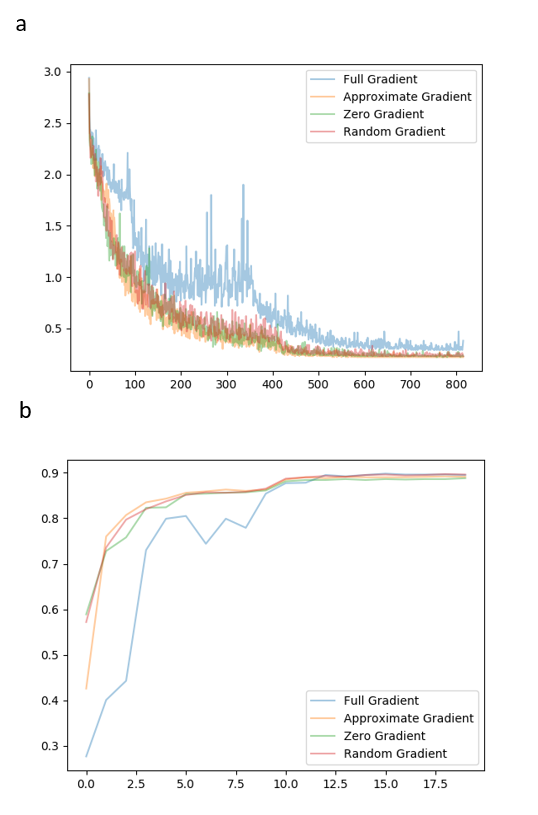}
\end{center}
   \caption{a) Training loss of VGG-19 model with different approximation methods. b) Validation accuracy of VGG-19 model with different approximation methods.}
\label{fig:vacc}
\end{figure}

For VGG-19, shown in Figure \ref{fig:vacc}, we see that full gradient descent actually lags that of the approximated methods in reaching target validation accuracy. In this case, all three approximation methods perform very well. However, full gradient descent does overtake all approximation methods finally in terms of validation accuracy. This suggests that perhaps a fruitful approach to explore, at least for networks similar to VGG, would be to use approximations early on in training and switch to full gradient computation when a validation accuracy plateau has been reached.

\subsection{Speedup-Accuracy Tradeoffs}

Here, we present the wall-clock speedups achieved for each network and approximation method. We compare the speedups against the validation accuracy loss, measured from the best validation accuracy achieved during training. Validation accuracy was calculated every ten epochs. As aforementioned, the random gradient implementation is quite inefficient and is pending future work. The speedup takes into account the overhead of defining a custom operation in Tensorflow, as well as the significant overhead of switching gradient computation on global training step. For the 2-layer CNN, we are unable to achieve wall-clock speedup for all approximation methods, even the zero gradient one, because of this overhead. (Table \ref{tab:2acc}) However, all approximation methods achieve little validation accuracy loss. The random gradient method even outperforms full gradient computation by 0.8\%. 

\begin{table}[t]
\begin{center}
\includegraphics[width=1\linewidth]{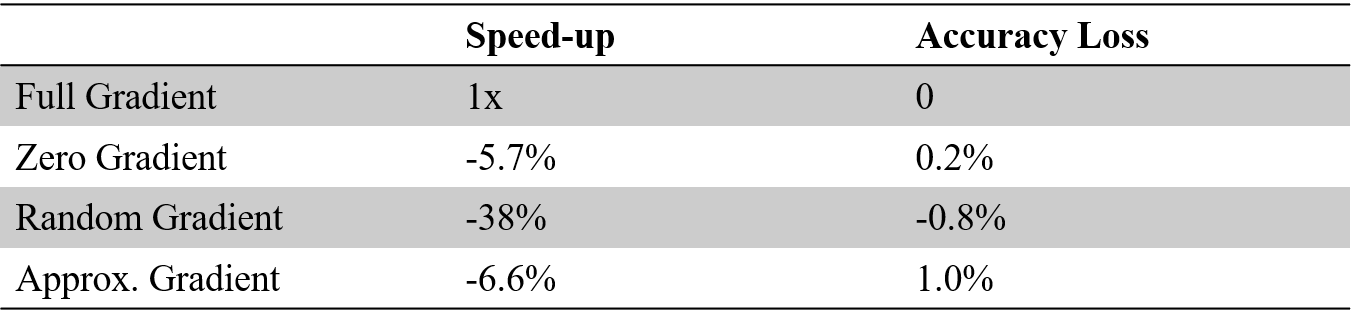}
\end{center}
   \caption{Training speedup and validation accuracy loss for the approximation methods on 2-layer CNN. Negative speedup indicates a slowdown.}
\label{tab:2acc}
\end{table}

For ResNet-20, the approximation schedule we choose does not involve switching gradient computations. We avoid the switching overhead and can achieve speedups for both the zero gradient method and the approximated gradient method. As shown in Table \ref{tab:racc}, the zero gradient method achieves roughly a third of the speedup compared to training the baseline ResNet-14 model. The approximated gradient method also achieves a 3.5\% wall-clock speedup, and is the only method to suffer less accuracy loss than just using a smaller ResNet-14. In the following section, we demonstrate that with other approximation schedules, the approximated gradient method can achieve as little as 0.1\% accuracy loss. 

\begin{table}[t]
\begin{center}
\includegraphics[width=1\linewidth]{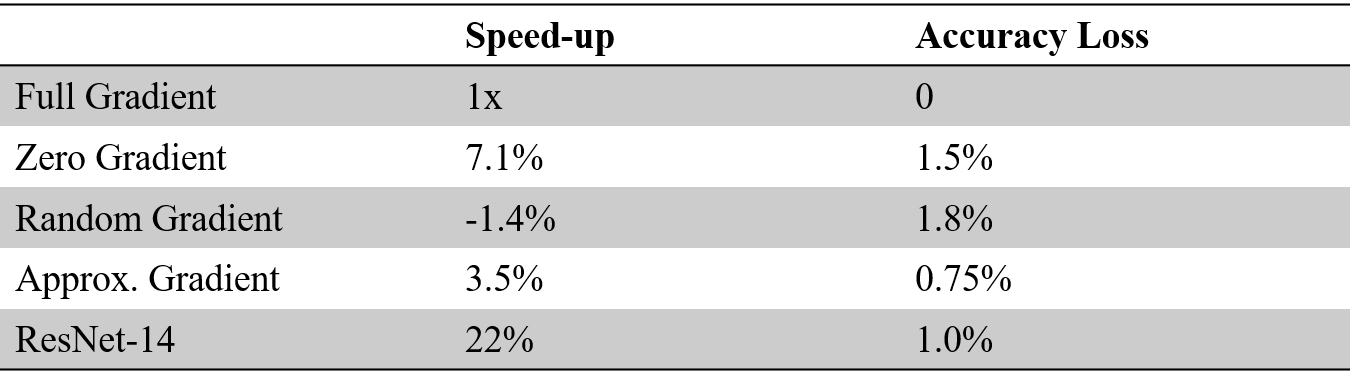}
\end{center}
   \caption{Training speedup and validation accuracy loss for the approximation methods on ResNet-20. Negative speedup indicates a slowdown.}
\label{tab:racc}
\end{table}

For VGG-19, despite being quicker to converge, the approximation methods all have worse validation accuracy than the baseline method. (Table \ref{tab:vacc}) The best approximation method appears to be the random gradient method, though it is extremely slow due to our inefficient implementation in Tensorflow. The other two methods also achieve high validation accuracies, with the approximated gradient method slightly better than the zero gradient method. Both methods are able to achieve speedups in training. 

\begin{table}[t]
\begin{center}
\includegraphics[width=1\linewidth]{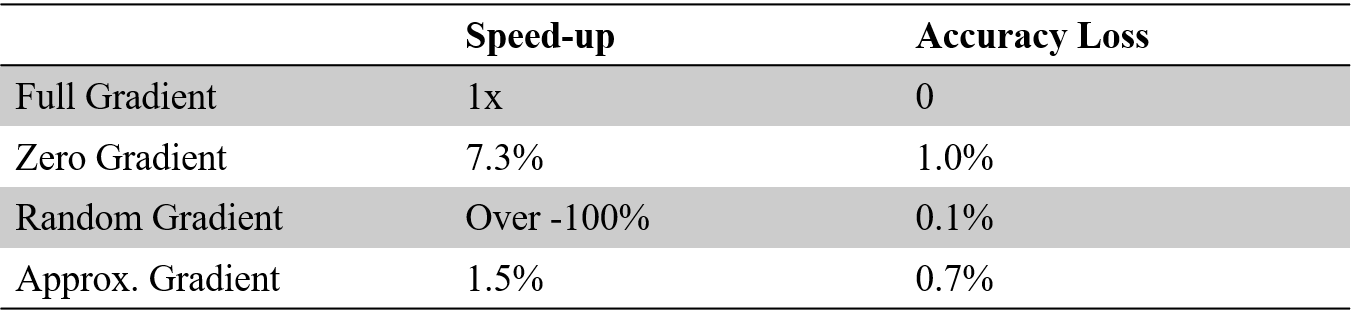}
\end{center}
   \caption{Training speedup and validation accuracy loss for the approximation methods on VGG-19. Negative speedup indicates a slowdown.}
\label{tab:vacc}
\end{table}

\subsection{Robustness to Approximation Schedule}

Here, we explore two new approximation schedules for ResNet-20, keeping the total proportion of the time we apply the approximation to 25 percent. We will refer to the approximation schedule presented in the secion above as schedule 1. Schedule 2 applies the selected approximation method every other layer for every other batch. Schedule 3 applies the selected approximation method every layer for every fourth batch. We also present the baseline result of the ResNet-14 model. 

\begin{figure}[t]
\begin{center}
\includegraphics[width=1\linewidth]{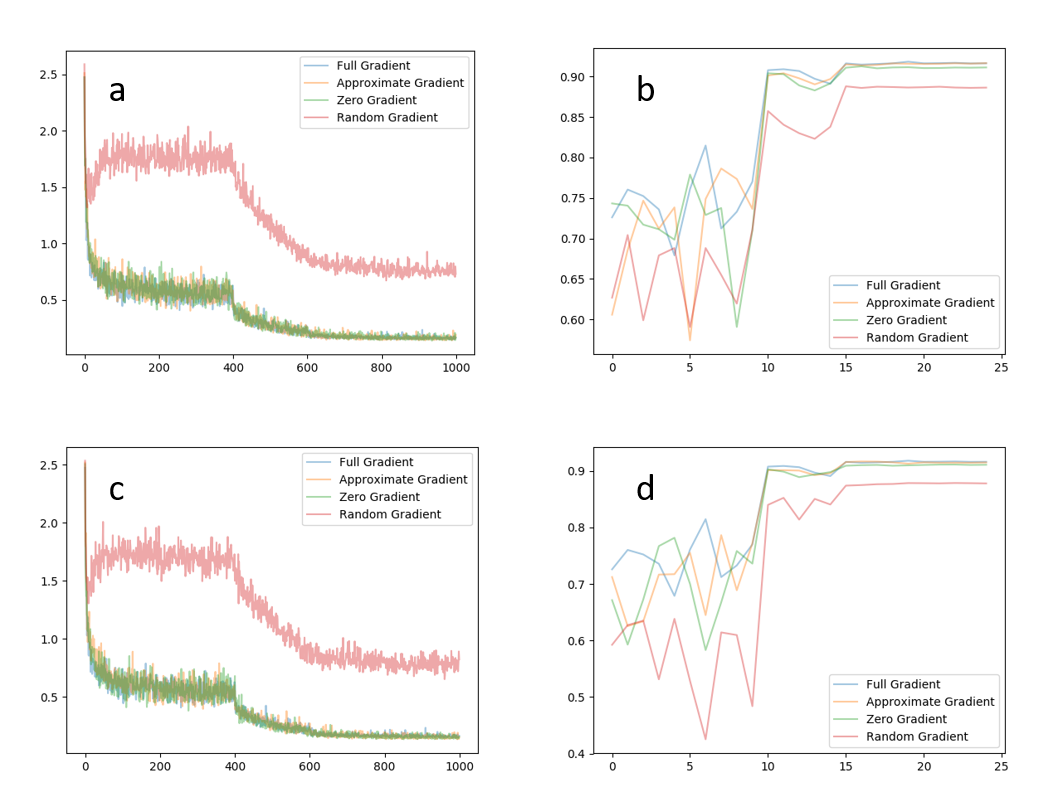}
\end{center}
   \caption{a) Training loss of ResNet-20 with different approximation methods for approximation schedule 2. b) Validation accuracy of ResNet-20 with different approximation methods for approximation schedule 2. c) Training loss of ResNet-20 with different approximation methods for approximation schedule 3. d) Validation accuracy of ResNet-20 with different approximation methods for approximation schedule 3.}
\label{fig:robust}
\end{figure}

As we can see from Figure \ref{fig:robust} and Table \ref{tab:robust-2}, under schedules 2 and 3, both the zero gradient and the approximated gradient method perform well. In fact, for the approximated gradient and the zero gradient methods the validation accuracy loss is smaller than schedule 1. Indeed, in schedule 3, the approximated gradient's best validation accuracy is within 0.1\% of that of the full gradient computation. The random gradient method's validation accuracy is in now line with its poor loss curve for these two approximation schedules. This suggests that the random gradient method does not work well for ResNet-20 architecture.

\begin{table}[t]
\begin{center}
\includegraphics[width=1\linewidth]{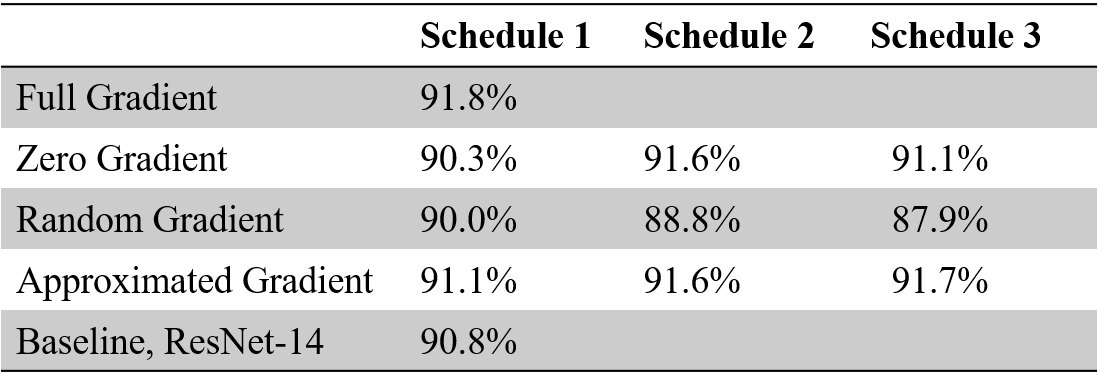}
\end{center}
   \caption{Validation accuracy for different approximation schedules on ResNet-20. Schedule 1 is the same as presented above.}
\label{tab:robust-2}
\end{table}

\section{Discussion and Conclusion}

While research on accelerating deep learning inference abounds, there is relatively limited work focused on accelerating the training process. Recent works such as PruneTrain prune the neural network in training, but suffers quite serious loss in validation accuracy \cite{lym2019prunetrain}. Approaches such as DropBack \cite{golub2018dropback} and MeProp \cite{wei2017minimal,sun2017meprop} show that approximated gradient are sufficient in successfully training neural networks but don't yet offer real wall-clock speedups. In this work, we show that we can train deep neural networks to good validation accuracy with very minimal gradient information on a subset of the layers, leading to wall-clock speedups for training. 

We are surprised by the consistent strong performance of the zero gradient method. For ResNet-20, for two of the three approximation schedules tested, the validation accuracy loss is better than that of a smaller baseline network. Its performance is also satisfactory on VGG-19 as well as the 2-layer CNN. It admits an extremely fast implementation that delivers consistent speedups. This points to a simple way to potentially boost training speed in deep neural networks, while maintaining their performance advantage over shallower alternatives. 

We also demonstrate that random gradient methods can train deep neural networks to convergence, provided they are only applied to a subset of the layers. For the 2-layer CNN and VGG-19, this method leads to the least validation accuracy loss of all three approximation methods. However, its performance serious lags other methods on ResNet-20, suggesting that its performance is network-architecture-specific. Naive feedback alignment, where the random gradient signal is fixed before training starts, has been shown to be difficult to extend to deep convolutional architectures  \cite{han2019efficient,bartunov2018assessing}.We show here that if the random gradients are newly generated every batch and applied to a subset of layers, they can be used to train deep neural networks to convergence. Interestingly, generating new random gradients every batch effectively abolishes any kind of possible ``alignment'' in the network, calling for a new explanation of why the network converges. Evidently, this method holds the potential for an extremely efficient implementation, something we are currently working on. 

Finally, we present a gradient approximation method with an efficient GPU implementation. Our approximation method is consistent in terms of validation accuracy across different network architectures and approximation schedules. Although the training wall clock time speedup isn't large, the validation accuracy loss is also small. We wish to re-emphasize here the small validation accuracy difference observed between the baseline ResNet-14 and ResNet-20, leading us to believe that novel training speed-up methods must incur minimal validation accuracy loss to be more practical than simply training a smaller network.

In conclusion, we show that we can ``fool" deep neural networks into training properly while supplying it only very minimal gradient information on select layers. The approximation methods are simple and robust, holding the promise to accelerate the lengthy training process for state-of-the-art deep CNNs.

\section{Future Work}
Besides those already mentioned, there are several more interesting directions of future work. One direction is predicting the validation accuracy loss that a neural network would suffer from a particular approximation schedule. With such a predictor, we can optimize for the fastest approximation schedule while constraining the final validation accuracy loss before the training run. We can also examine the effects of mingling different approximation methods and integrating existing methods such as PruneTrain and Dropback \cite{lym2019prunetrain, golub2018dropback}. Another direction is approximating  the gradient of the hidden activations, as is done in meProp \cite{sun2017meprop}. However, if we approximate the hidden activations at a deeper layer of the network, the approximation error will be propagated to the shallower layers. Due to this concern, we start with approximating filter weight gradients, where the effect of errors are local. Finally, we are working on integrating this approach into a distributed training setting, where the approximation schedule is now 3-dimensional (machine, layer, batch). This approach would be crucial for the approximation methods to work with larger scale datasets such as ImageNet, thus potentially allowing for wall-clock speed-up in large scale training.

{\small
\bibliographystyle{ieee_fullname}
\bibliography{mypaper}
}

\end{document}